\documentclass[runningheads]{llncs}

\usepackage{eccvabbrv}

\usepackage{graphicx}
\usepackage{booktabs}
\usepackage{amsmath}
\usepackage{multicol}
\usepackage{multirow}
\usepackage{algorithm}
\usepackage{algorithmic}
\usepackage{amsmath} 
\usepackage{amsfonts}
\usepackage{subfigure}
\usepackage{subcaption} 

\usepackage[accsupp]{axessibility}  

\usepackage{hyperref}

\usepackage{orcidlink}

\begin{document}

\title{Beyond Semantics: Uncovering the Physics of Fakes via Universal Physical Descriptors for Cross-Modal Synthetic Detection} 


\author{Mei Qiu\inst{1,2} \and
Jianqiang Zhao\inst{1}\and
Yanyun Qu\inst{2}}


\institute{SDIC Intelligence Information Technology Co., Ltd., \and Xia'men University, China\\
\email{qiu172@purdue.edu, meiya.ai.1@outlook.com, yyqu@xmu.edu.cn}\\
}

\maketitle

\begin{abstract}
The rapid advancement of AI-generated content (AIGC) has blurred the boundaries between real and synthetic images, exposing the limitations of existing deepfake detectors that often overfit to specific generative models. This adaptability crisis calls for a fundamental re-examination of the intrinsic physical characteristics that distinguish natural from AI-generated images. In this paper, we address two critical research questions: (1) What physical features can stably and robustly discriminate AI-generated images across diverse datasets and generative architectures? (2) Can these objective pixel-level features be integrated into multi-modal models like CLIP to enhance detection performance while mitigating the unreliability of language-based information? To answer these questions, we conduct a comprehensive exploration of 15 physical features across more than 20 datasets generated by various GANs and diffusion models. We propose a novel feature selection algorithm that identifies five core physical features—including Laplacian variance, Sobel statistics, and residual noise variance—that exhibit consistent discriminative power across all tested datasets. These features are then converted into text-encoded values and integrated with semantic captions to guide image-text representation learning in CLIP. Extensive experiments demonstrate that our method achieves state-of-the-art performance on multiple Genimage benchmarks, with near-perfect accuracy (99.8\%) on datasets such as Wukong and SDv1.4. By bridging pixel-level authenticity with semantic understanding, this work pioneers the use of physically grounded features for trustworthy vision-language modeling and opens new directions for mitigating hallucinations and textual inaccuracies in large multi-modal models.
\end{abstract}

\section{Introduction}
\label{sec:intro}
The rapid advancement of AIGC technology \cite{ho2020denoising,rombach2022high,team2023gemini} has increasingly blurred the boundary between real and synthetic images, posing unprecedented challenges to authenticity detection \cite{lin2024detecting,nguyen2025deepfake} and raising significant societal risks \cite{saetra2023generative,shao2023detecting}. Traditional data-driven methods, which rely heavily on pre-trained models and specific dataset distributions, often struggle to generalize across evolving generative models \cite{dzanic2020fourier,tan2024rethinking}. This adaptability crisis calls for a fundamental re-examination of intrinsic physical characteristics that stably and robustly distinguish AI-generated from natural images \cite{zhou2025beyond,theis2024makes,zhang2025physics}. Meanwhile, multi-modal models like CLIP show promise in deepfake detection by leveraging semantic information, yet the reliability of language—especially when LLM-generated—remains questionable \cite{ojha2023towards,tan2025c2p}. In contrast, pixel-level information derived directly from images remains objective and unalterable, motivating our core insight: ``Pixels Don't Lie'' \cite{kuckreja2026pixels}.

To address these challenges, we investigate two critical research questions. \textbf{First}, what physical features can universally and robustly discriminate AI-generated from natural images across diverse generative models? \textbf{Second}, can these objective pixel-level features be integrated into multi-modal models like CLIP to enhance detection performance while mitigating the unreliability of language-based information? This study makes two key contributions. (\textit{i}) We comprehensively explore 15 physical features—including Laplacian variance, Sobel statistics, DCT variance, and residual noise—across over 20 datasets from various GANs and diffusion models. We propose a novel algorithm to evaluate feature stability and discriminability, ultimately identifying fou e or five core features with consistent cross-dataset performance. (\textit{ii}) We convert these core features into text-encoded values and merge them with semantic captions to enhance embedding extraction and deepfake detection in CLIP. Beyond improving accuracy, this integration bridges pixel-level authenticity and semantic understanding, offering new directions for building trustworthy vision-language models.

\section{Related Works}
We review existing literature on synthetic detection (including face forgery and general AIGC detection) by categorizing methodologies into two principal domains: data-driven and physics-based approaches.

\textbf{Data-driven methods} dominate the field, leveraging deep learning to distinguish real from synthetic content. Early work by Rossler et al.~\cite{rossler2019faceforensics++} employed Xception for face manipulation detection, while Wang et al.~\cite{wang2021representative} addressed limited model focus via attention-based augmentation. To improve generalization, Cao et al.~\cite{cao2022end} proposed a reconstruction-classification framework, Chen et al.~\cite{chen2022self} utilized augmented forgeries with adversarial training, and Shiohara and Yamasaki~\cite{shiohara2022detecting} introduced SBIs for robust representations. With the advent of foundation models, Ojha et al.~\cite{ojha2023towards} leveraged pre-trained vision-language features in a non-learning approach, Liu et al.~\cite{liu2024forgery} developed FatFormer to integrate frequency features, and Tan et al.~\cite{tan2025c2p} proposed C2P-CLIP for detection. Recent advances include disentanglement frameworks~\cite{yan2023ucf}, latent space augmentation~\cite{yan2024transcending}, and fusion of co-occurrence matrices with CNNs~\cite{nataraj2019detecting}. This evolution reflects the field's progression from CNN-based artifact mining to modern pre-trained and disentanglement approaches.

\textbf{Physics-based and physics-informed methods} leverage inherent physical constraints—such as frequency spectra, spatial texture, gradients, and temporal dynamics—to identify stable synthetic traces. Frank et al.~\cite{frank2020leveraging} showed GAN upsampling introduces consistent frequency artifacts, while Chai et al.~\cite{chai2020makes} found local patch artifacts persist even in refined generators. Liu et al.~\cite{liu2020global} used global texture statistics via Gram-Net, and Haliassos et al.~\cite{haliassos2021lips} proposed LipForensics targeting lip movement dynamics. Subsequent work introduced re-synthesis frameworks~\cite{he2021beyond}, frequency filtering~\cite{jeong2022bihpf}, adaptive texture-frequency mining~\cite{liu2023adaptive}, and gradient-based representations~\cite{tan2023learning}. Tan et al.~\cite{tan2024rethinking} improved generalization by modeling upsampling inconsistencies, while Cheng et al.~\cite{cheng2025co} fused pixel artifacts with semantic anomalies. For video detection, Zhang et al.~\cite{zhang2025physics} proposed a normalized spatiotemporal gradient (NSG) quantifying deviations from natural dynamics. Unlike data-driven methods, physics-based approaches offer stronger robustness and generalization by relying on fundamental constraints rather than dataset-specific patterns, though they often require domain expertise to design effective features.
\begin{figure}[tb]
  \centering
  \includegraphics[height=6.5cm]{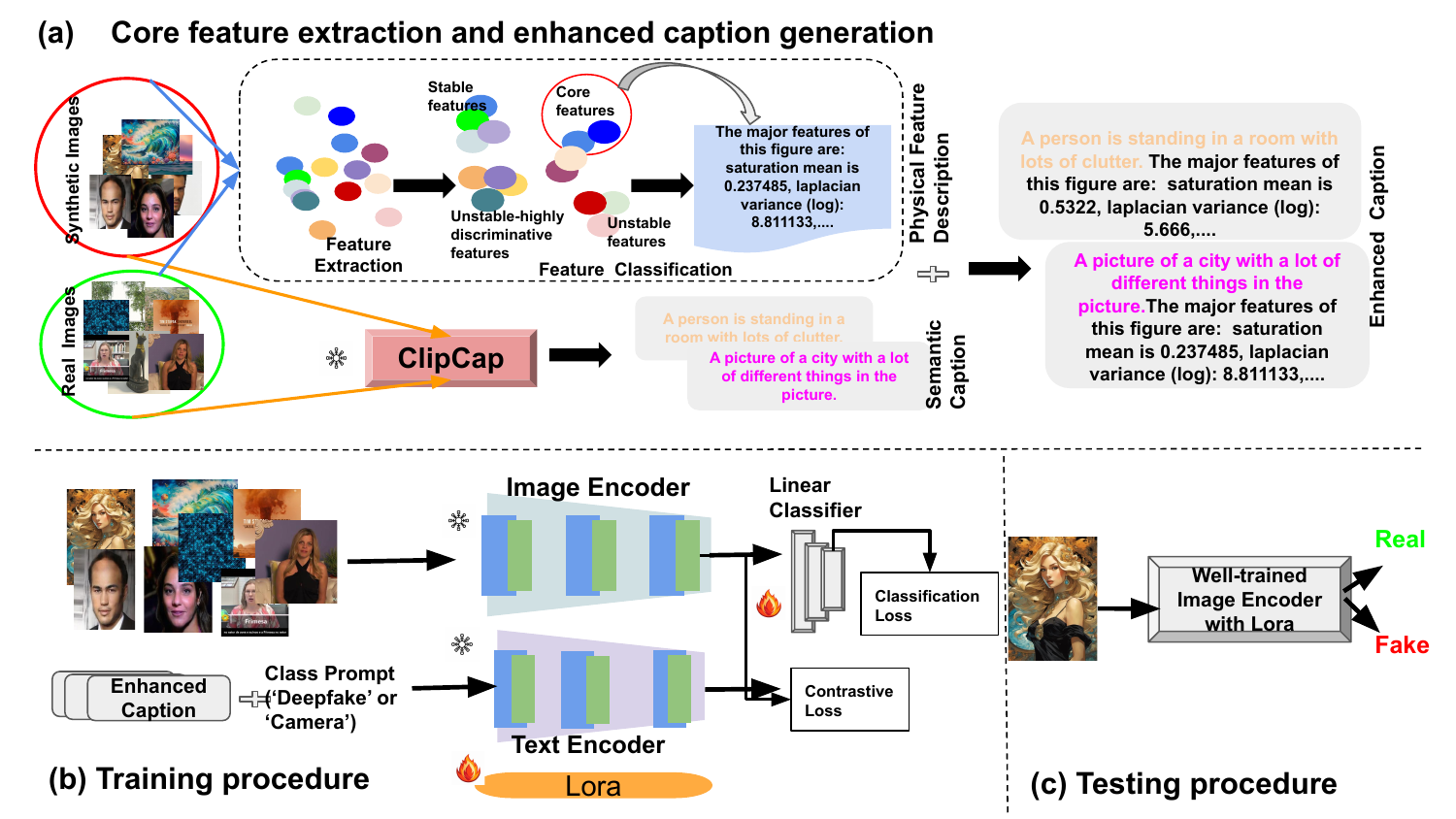}
  \caption{\small Overall workflow of the proposed synthetic image detection framework. (a) Enhanced caption preparation: Integrate input images’ core physical features into captions to enrich representation. (b) Training: Use enhanced captions and class prompts to train the model for real/AI-generated image discrimination. (c) Testing: Input images into the trained model to predict AI generation. The training and testing procedure are totally the same with C2pClip\cite{tan2025c2p}.}
  \label{fig:whole_frame}
\end{figure}

\section{Methods}
Inspired by FakeBench \cite{li2025fakebench}, Explainable MLLM \cite{ji2025towards}, and C2P-CLIP \cite{tan2025c2p}, we propose and comprehensively evaluate 15 classic physical features. These 15 discriminative visual attributes are computed to quantify statistical and perceptual discrepancies between authentic and AI-generated imagery. Each feature targets distinct artifacts inherent in synthetic content—such as unnatural smoothness, color inconsistency, or repetitive patterns—with mathematical formulations tailored to amplify these differences and normalize feature ranges for robust model input. We design a new algorithm to estimate the stability and discriminative power of each feature. We then select core features that exhibit both high stability and strong discrimination and incorporate these features into text descriptions. The enhanced captions are formed by merging the original ClipCap \cite{mokady2021clipcap} generated texts with these new core feature-based descriptions. We adopt the training and testing pipeline from C2P-CLIP to investigate whether core features-based descriptions can improve feature understanding and benefit fake image detection. The overall framework is illustrated in Fig. \ref{fig:whole_frame}.
\subsection{Feature Definition and Extraction}
Each feature is normalized or log-scaled to stabilize its dynamic range. The 15 features are formally defined as in Table \ref{tab:features_min_space}: 
\begin{table*}[tb]
\centering
\fontsize{7}{8}\selectfont 
\setlength{\tabcolsep}{0pt} 
\renewcommand{\arraystretch}{0.15} 
\setlength{\mathsurround}{0pt} 
\begin{tabular}{@{}ccc@{}} 
\toprule
$F_1\!=\!\frac{1}{HW}\sum_{i,j}\text{HSV}_{i,j,1}$ &
$F_2\!=\!\frac{1}{HW}\sum_{i,j}\text{HSV}_{i,j,2}$ &
$F_3\!=\!\log\!\left(1+\operatorname{Var}(\nabla^2 I_g)\right)$ \\
\midrule
$F_4\!=\!10\!\cdot\!\frac{1}{HW}\sum_{i,j}\sqrt{S_x^2\!+\!S_y^2}$ &
$F_5\!=\!10\!\cdot\!\sqrt{\frac{1}{HW}\sum_{i,j}(M_{i,j}\!-\!\mu_M)^2}$ &
$F_6\!=\!\log\!\left(1+\frac{1}{N}\sum_k\operatorname{Var}(\operatorname{DCT}(B_k))\right)$ \\
\midrule
$F_7\!=\!\log\!\left(1+1000\operatorname{Var}(R)\right)$ &
$F_8\!=\!-\sum_b p_b\log_2 p_b$ &
$F_9\!=\!\frac{\operatorname{Cov}(R,G)}{\sqrt{\operatorname{Var}(R)\operatorname{Var}(G)}}$ \\
\midrule
$F_{10}\!=\!\frac{\operatorname{Cov}(R,B)}{\sqrt{\operatorname{Var}(R)\operatorname{Var}(B)}}$ &
$F_{11}\!=\!-\sum_v p_v\log_2 p_v$ &
$F_{12}\!=\!-\sum_v p_v\log_2 p_v$ \\
\midrule
$F_{13}\!=\!\log\!\left(1+\operatorname{Var}(\text{HSV}_{:,:,0})\right)$ &
$F_{14}\!=\!\frac{1}{HW}\sum_{i,j}\frac{E_{i,j}}{255}$ &
$F_{15}\!=\!\log\!\left(1+|\operatorname{Kurt}(B)|\right)$ \\
\bottomrule
\end{tabular}
\vspace{-1pt}
\caption{\tiny Mathematical definitions of 15 synthetic image detection features. Key symbol definitions: $i,j$ = 2D pixel coordinates (height/width); $H,W$ = image height/width; $\text{HSV}_{i,j,c}$ = HSV color space value at pixel $(i,j)$ for channel $c$ (1=saturation, 2=brightness, 0=hue); $I_g$ = grayscale image; $\nabla^2$ = Laplacian operator (second-order derivative for sharpness); $S_x/S_y$ = Sobel horizontal/vertical edge responses; $M_{i,j}$ = Sobel edge magnitude at $(i,j)$; $\mu_M$ = mean edge magnitude; $\operatorname{DCT}(\cdot)$ = Discrete Cosine Transform; $B_k$ = $k$-th 8×8 image block; $N=1000$ (sampled blocks for efficiency); $R$ = residual noise (image minus non-local means denoised version); $p_b$ = normalized frequency of Local Binary Pattern (LBP) bin $b$; $\operatorname{Cov}(\cdot,\cdot)$ = covariance; $\operatorname{Var}(\cdot)$ = variance; $R/G/B$ = red/green/blue color channels; $p_v$ = normalized pixel value frequency for chrominance channels (Cr/Cb); $E_{i,j}$ = Canny edge detector output (0=non-edge, 255=edge); $\operatorname{Kurt}(\cdot)$ = kurtosis (tailedness of intensity distribution).}
\label{tab:features_min_space}
\end{table*}
\begin{align}
\text{CV}(f) &= \frac{\sigma_{\mu_f}}{\mu_{\mu_f}}, \label{eq:cv} \\
S_s(f) &= \max\left(0, \min\left(1, 1 - \text{CV}(f)\right)\right), \label{eq:stability} \\
\text{JMD}(f) &= \frac{|\mu_{\text{real}}(f) - \mu_{\text{fake}}(f)|}{\sigma_{\text{real}}(f) + \sigma_{\text{fake}}(f)}, \label{eq:jmd} \\
S_d(f) &= \max\left(0, \min\left(1, \frac{\overline{\text{JMD}}(f) + \overline{\text{AUC}}(f)}{2}\right)\right), \label{eq:discriminability}
\end{align}
\small where:
$\sigma_{\mu_f}$ = cross-dataset standard deviation of feature $f$'s dataset-wise means,
$\mu_{\mu_f}$ = mean of feature $f$'s dataset-wise means,
$\mu_{\text{real}}(f)/\mu_{\text{fake}}(f)$ = mean of feature $f$ for real/fake samples,
$\sigma_{\text{real}}(f)/\sigma_{\text{fake}}(f)$ = standard deviation of feature $f$ for real/fake samples,
$\overline{\text{JMD}}(f)$ = mean JMD of feature $f$ across datasets,
$\overline{\text{AUC}}(f)$ = mean AUC of feature $f$ across datasets (via logistic regression).

\begin{algorithm}[tb]
\small 
\caption{FSDVA (Feature Stability-Discriminability Validation Assessment)}
\label{alg:fsdva}
\begin{algorithmic}[1] 
\REQUIRE $D$: Dictionary of dataset-wise features $\{\mathcal{D}_k: \{\mathbf{F}_{\text{real}}^k \in \mathbb{R}^{N \times 15}, \mathbf{F}_{\text{fake}}^k \in \mathbb{R}^{N \times 15}\}\}$ (for datasets $\mathcal{D}_k$)
\ENSURE $\mathcal{M}$: Metric table with $S_s(f)$, $S_d(f)$, and class for each feature $f$

\FOR{each feature $f \in \{f_1, f_2, ..., f_{15}\}$}
    \STATE \textbf{Step 1: Compute Stability Score } $S_s(f)$ \label{step:stability}
    \STATE For each dataset $\mathcal{D}_k$: compute mean $\mu_f^k$ of $f$ (merge $\mathbf{F}_{\text{real}}^k[:,f]$ and $\mathbf{F}_{\text{fake}}^k[:,f]$)
    \STATE Compute cross-dataset coefficient of variation $\text{CV}(f)$ (Eq. \ref{eq:cv})
    \STATE $S_s(f) = \max\left(0, \min\left(1, 1 - \text{CV}(f)\right)\right)$ (Eq. \ref{eq:stability})

    \STATE \textbf{Step 2: Compute Discriminability Score } $S_d(f)$ \label{step:discriminability}
    \STATE For each dataset $\mathcal{D}_k$:
        \STATE Compute inter/intra-class distance ratio $\text{JMD}(f)$ (Eq. \ref{eq:jmd}) for $f$ (real vs. fake)
        \STATE Train logistic regression on $f$ (real/fake labels) and compute AUC
    \STATE Compute mean JMD/AUC across datasets: $\overline{\text{JMD}}(f)$, $\overline{\text{AUC}}(f)$
    \STATE $S_d(f) = \max\left(0, \min\left(1, \frac{\overline{\text{JMD}}(f) + \overline{\text{AUC}}(f)}{2}\right)\right)$ (Eq. \ref{eq:discriminability})

    \STATE \textbf{Step 3: Assign Feature Class} \label{step:classification}
    \STATE \quad $\text{Class}(f) = \begin{cases}
        \text{Core Feature} & S_s(f) \geq 0.7 \land S_d(f) \geq 0.5, \\
        \text{Usable Feature} & S_s(f) \geq 0.45 \land S_d(f) \geq 0.3, \\
        \text{Unstable High-Discrim} & S_s(f) < 0.7 \land S_d(f) \geq 0.4, \\
        \text{Unusable Feature} & \text{otherwise}.
    \end{cases}$

    \STATE Append $\{f, S_s(f), S_d(f), \text{Class}(f)\}$ to $\mathcal{M}$
\ENDFOR
\RETURN $\mathcal{M}$
\end{algorithmic}
\label{algo:FSDVA}
\end{algorithm}
Specificly, these 15 features are: saturation mean ($F_1$), brightness mean ($F_2$), laplacian variance ($F_3$), sobel magnitude mean ($F_4$), 
sobel magnitude std ($F_5$), DCT variance ($F_6$), residual noise variance ($F_7$), lbp entropy  ($F_8$), rg correlation ($F_9$), rb correlation ($F_{10}$), chroma entropy dim-1 ($F_{11}$), chroma entropy dim-2 ($F_{12}$), hue variance ($F_{13}$), canny edge density ($F_{14}$), and blue channel kurtosis (($F_{15}$)).
\subsection{Feature's Stability and Discrimativity Evaluation}
To comprehensively evaluate the effectiveness of the 15 handcrafted features for synthetic image detection, we propose the Feature Stability-Discriminability Validation Assessment (FSDVA) algorithm \ref{algo:FSDVA}. This algorithm systematically quantifies two key metrics for each feature across multiple datasets: cross-dataset stability and real-fake discriminability. For stability assessment, FSDVA first computes the coefficient of variation of each feature’s dataset-wise means using Eq. \ref{eq:cv}, then converts this coefficient into a normalized stability score via Eq. \ref{eq:stability} to reflect the feature’s consistency across different data distributions. For discriminability assessment, it first calculates the inter-class and intra-class distance ratio (JMD) for each feature using Eq. \ref{eq:jmd}, integrates this JMD with the AUC score from logistic regression classifiers (trained to distinguish real from fake images) across datasets, and finally yields a normalized discriminability score through Eq. \ref{eq:discriminability}. Finally, FSDVA classifies each feature into one of four categories based on its stability and discriminability scores, enabling the selection of optimal features for robust synthetic image detection across diverse datasets.

\begin{figure}[tb]
  \centering
\includegraphics[width=0.9\textwidth, height=6.5cm, keepaspectratio]{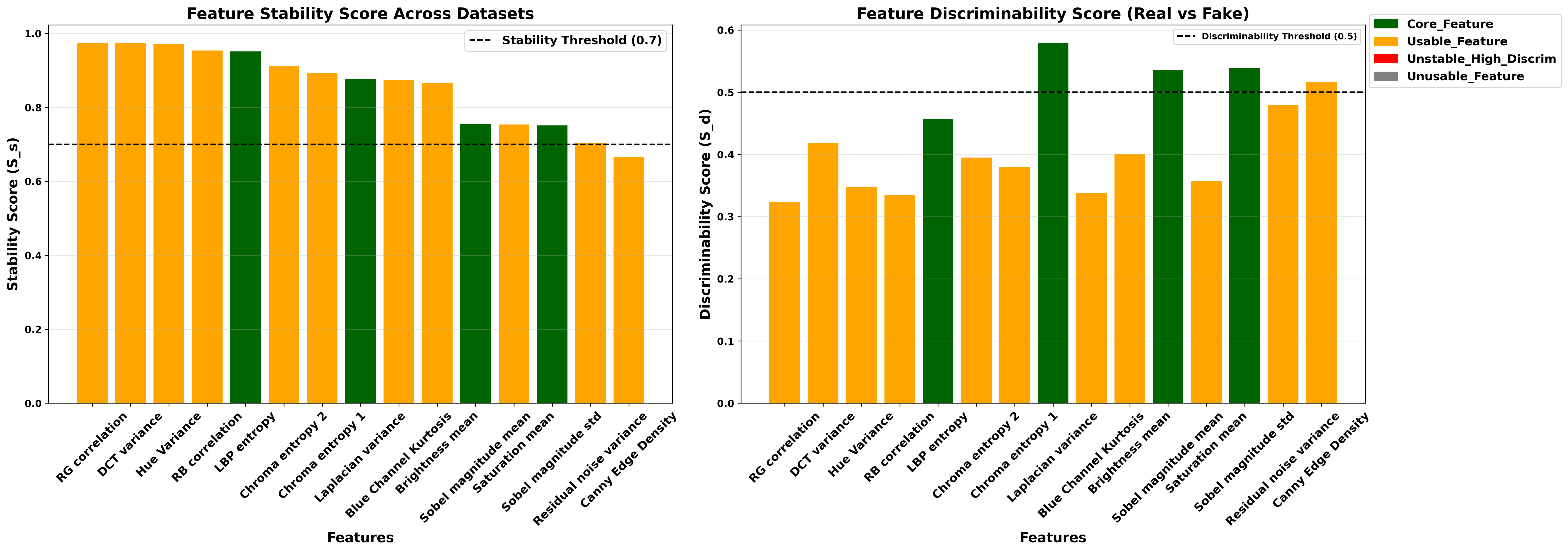}
  \caption{\small Stability ($S_s$, left) and discriminability ($S_d$, right) scores of image features for fake image detection. Based on the thresholds, set features to Core Feature(green), Usable Feature(orange), red Unstable High-Discrim(red), and Unusable Feature(grey).}
  \label{fig:feat_cls}
\end{figure}
\begin{figure}[tb]
  \centering
 \includegraphics[width=1.2\textwidth, height=6.5cm, keepaspectratio]{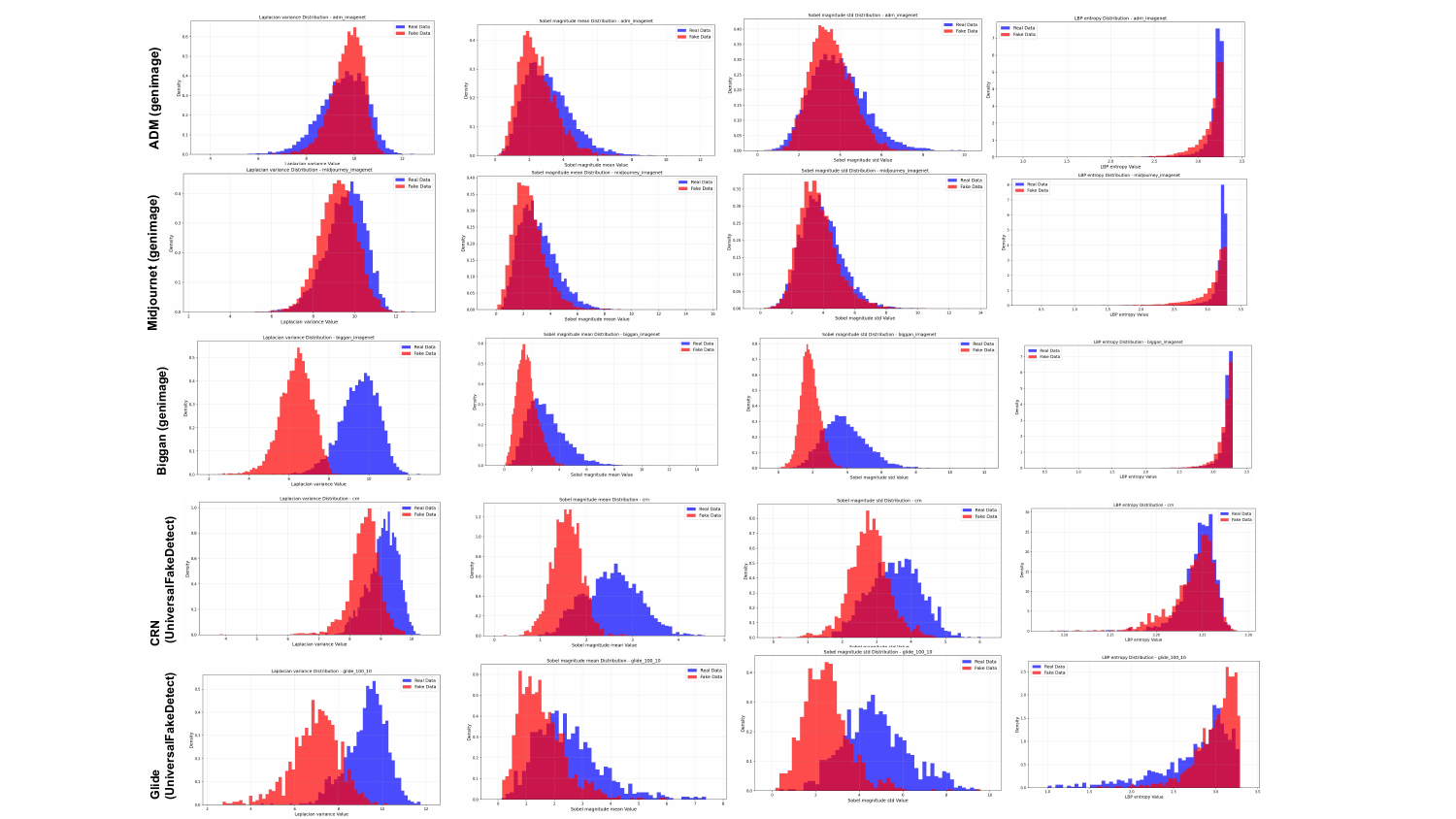}
  \caption{\small Density distributions of four core features (Laplacian variance, Sobel magnitude mean/std, LBP variance) over ADM/Midjourney (GenImage) and BigGAN/CRN/Glide (UniversalFakeDetect). Blue = real images, red = fake images.}
  \label{fig:core_feat}
\end{figure}
\begin{figure}[tb]
  \centering
\includegraphics[width=0.9\textwidth, height=6.5cm, keepaspectratio]{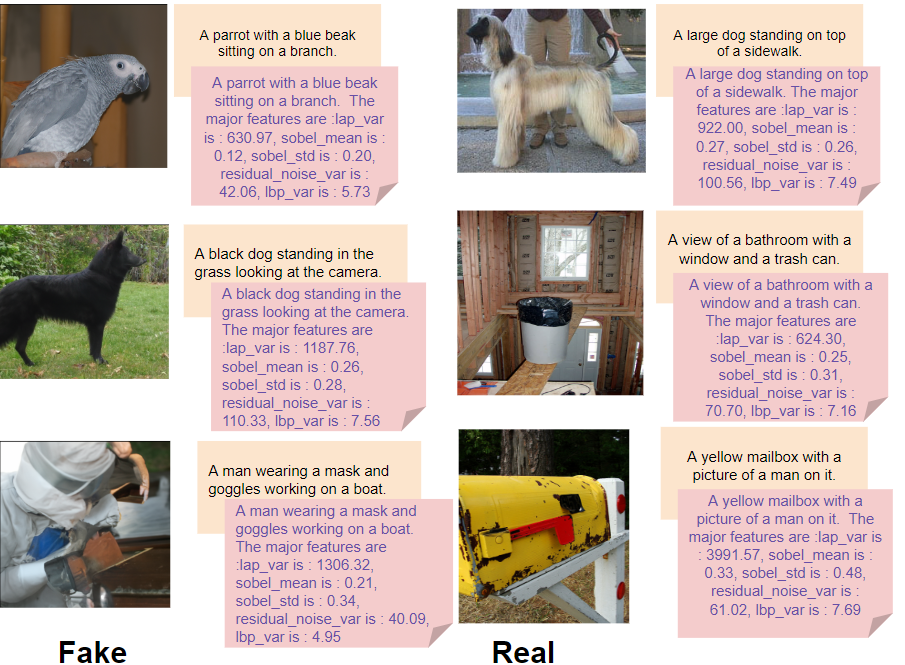}
  \caption{\small Image-text pairs with original texts generated by ClipCap and enhanced by core features' description.}
  \label{fig:cap_samples}
\end{figure}
\begin{table*}[t]
\centering
\resizebox{\linewidth}{!}{
\begin{tabular}{llcccccccccc}
\toprule
\multirow{2}{*}{Method} & \multirow{2}{*}{Training Dataset} & \multicolumn{8}{c}{Test Datasets (Genimage)} & \multirow{2}{*}{mAcc} \\
\cline{3-10}
& & Midjourney & SDv1.4 & SDv1.5 & ADM & GLIDE & Wukong & VQDM & BigGAN & \\
\midrule
\multirow{6}{*}{C2pClip} 
& Midjourney &97.2 &96.5  &96.2 & 79.8 &96.6  &93.6  &87.0  &92.7  & 92.4 \\
& SDv1.4 & 88.2 & 90.9 & 97.9 & 96.4 & 99.0 & 98.8 & 96.5 & 98.7 & 95.8 \\
& SDv1.5 & 96.2 & 98.8 & 98.5 &92.8  & 98.5 &98.4  & 97 & 97.4 &97.2  \\
& ADM &86.4 & 96.8& 96.4 &98.4  &98.6  &95.1  & 98.2 &98.2  & 96.0 \\
& wukong &96.4  &98.9  & 98.4 & 92.9&98.4 & 98.5 & 97.9 & 97.2 & 97.3 \\
& vqdm & 94.1 & 97.0 &96.9 &96.7 &97.1 & 97.0 &97.3  &97.1 & 96.6 \\
\midrule
\multirow{6}{*}{Ours} 
& Midjourney &94.7  & \textbf{98.2} &\textbf{98.1}  & \textbf{82.5} & 94.1&93.5  &\textbf{88.7}  &\textbf{93.3}  & \textbf{92.9} \\
& SDv1.4 & \textbf{96.4} &\textbf{97} &96.7  &92.4 & 97 & 96.7 &96.1  &96.0 & \textbf{96.0} \\
& Sdv1.5 &95.4  & \textbf{99.0} & \textbf{98.7} &\textbf{94.3} & 97.0 & 98.4&95.6  &95.3 &96.7  \\
& ADM & \textbf{91.2} &\textbf{97.3} &\textbf{97.0}  &\textbf{99.8 } & 95.7 & 93.1& \textbf{98.9} &95.8 & \textbf{96.1} \\
& wukong &95.8  &\textbf{99.8}  & \textbf{99.6} & \textbf{93.1} &\textbf{99.2}  &\textbf{99.7}  &\textbf{99.2} & 96.2 &\textbf{97.8}  \\
& vqdm &\textbf{94.6} &\textbf{99.8} &\textbf{99.6}  &94.4  & 96.8 &\textbf{99.5}  & \textbf{99.9}&96.1 & \textbf{97.6} \\
\bottomrule
\end{tabular}
}
\caption{Impact of Training Dataset on Test Accuracy (Genimage Dataset). All methods are evaluated on the same 8 fixed test datasets (Midjourney, SDv1.4, SDv1.5, ADM, GLIDE, Wukong, VQDM, BigGAN) but trained on distinct datasets (Midjourney, SDv1.4, SDv1.5). Results highlight how training data choice affects cross-model generalization for C2pClip and our method (Ours), with \textcolor{blue}{bold values indicating our method achieve better performance on the corresponding test dataset}.}
\label{tab:genimage_diff_train}
\end{table*}
\begin{table*}[t]
\centering
\resizebox{\linewidth}{!}{
\begin{tabular}{lcccccccccc}
\toprule
Method & Ref & Midjourney & SDv1.4 & SDv1.5 & ADM & GLIDE & Wukong & VQDM & BigGAN & mAcc \\
\midrule
UniFD(2023) & CVPR2023 & 93.9 & 96.4 & 96.2 & 71.9 & 85.4 & 94.3 & 81.6 & 90.5 & 88.8 \\
NPR (2024c) & CVPR2024 & 81.0 & 98.2 & 97.9 & 76.9 & 89.8 & 96.9 & 84.1 & 84.2 & 88.6 \\
FreqNet(2024b) & AAAI2024 & 89.6 & 98.8 & 98.6 & 66.8 & 86.5 & 87.3 & 75.8 & 81.4 & 86.8 \\
FatFormer(2024) & CVPR2024 & 92.7 & 100.0 & 99.9 & 75.9 & 88.0 & 99.9 & 98.8 & 55.8 & 88.9 \\
C2pClip (Trump.Biden) &AAAI2025  & 82.2 & 95.1 & 95.5 & 95.1 & 98.9 & 98.7 & 93.8 & 98.3 &94.7 \\
C2pClip (Deepfake.Camera) &AAAI2025  & 88.2 & 90.9 & 97.9 & 96.4 & 99.8 & 98.8 & 96.5 & 98.7 & 95.8 \\
Ours (Deepfake.Camera) &  & \textbf{96.4} &97 &96.7  &92.4 & 97 & 96.7 &96.1  &96.0 & \textbf{96.0} \\
\bottomrule
\end{tabular}
}
\caption{Cross-model Accuracy (Acc) Performance on the Genimage Dataset. 
}
\label{tab:genimage_acc}
\end{table*}
\begin{figure}[tb]
  \centering
 \includegraphics[width=1.2\textwidth, height=6.5cm, keepaspectratio]{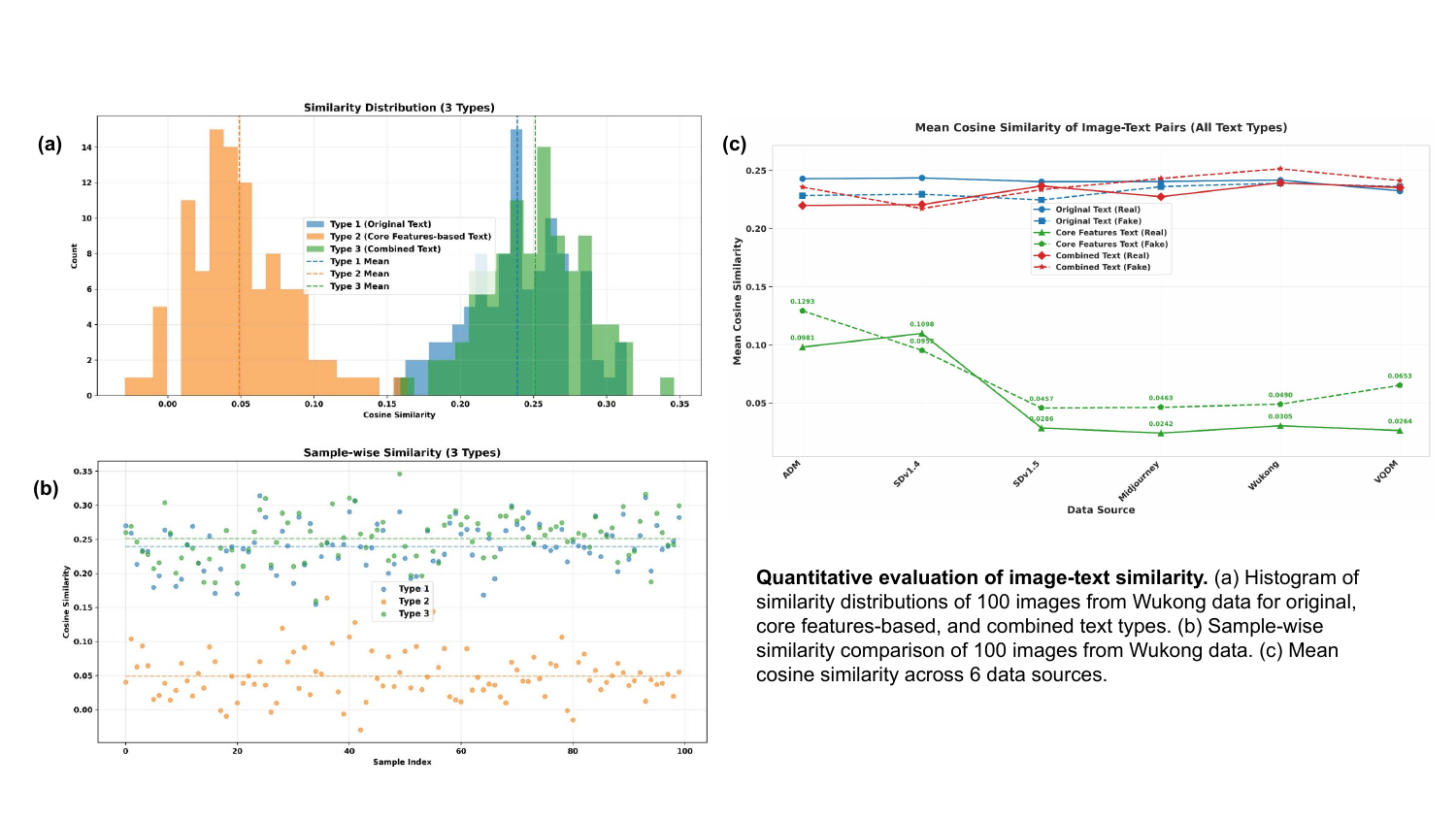}
  \caption{\small Quantitative evaluation of image-text similarity based on pre-trained Clip model.}
  \label{fig:Quantitative_evaluation}
\end{figure}
\subsection{Train Clip with Enhanced Caption by Merging Core Features}
As shown in Fig.~\ref{fig:whole_frame}, our method follows the same training and testing pipeline as C2pClip~\cite{tan2025c2p}, with the only modification being the enrichment of input captions with core physical features extracted via FSDVA (Algorithm.~\ref{algo:FSDVA}). 

Given a training set \( X = \{x_j, y_j\}_{j=1}^N \) with \( y \in \{0,1\} \) indicating fake (1) or real (0), each image \( x_j \) has an associated caption \( c_j \) generated by ClipCap. We compute a core feature set \( \mathcal{F}(x_j) \) and convert it into a descriptive text \( f_j \). The enhanced caption is formed by concatenation: \( \widetilde{c}_j = c_j \oplus f_j \). 

Text and image features are extracted via \( u_j = \text{encoder}_{\text{text}}(\widetilde{c}_j) \) and \( v_j = \text{encoder}^{\text{Lora}}_{\text{img}}(x_j) \). The training objective combines contrastive and classification losses:
\begin{equation}
   \mathcal{L} = \mathcal{L}_{\text{contrastive}} + \alpha \cdot \mathcal{L}_{\text{classification}}, 
\end{equation}
where \( \mathcal{L}_{\text{contrastive}} \) is the symmetric cross-entropy loss between paired image-text embeddings. At test time, only the image encoder with LoRA and the classifier are used for detection:
\begin{equation}
   p = \text{classifier}\left( \text{encoder}^{\text{Lora}}_{\text{img}}(x) \right). 
\end{equation}

\section{Experiments}
In this section, we provide the evaluation encompasses various aspects such as datasets, implementation details, and deepfake synthetic performance.
\subsection{Experiments Settings}
\vspace{-2mm}
\smallskip
\noindent
\textbf{Data.} Our feature extraction data is randomly sampled from the test subsets of GenImage~\cite{zhu2023genimage} and UniversalFakeDetect~\cite{ojha2023towards}. GenImage includes images from Midjourney, SDv1.4, SDv1.5, ADM, GLIDE, Wukong, VQDM, and BigGAN. From UniversalFakeDetect, we select 10 subsets: BigGAN, StarGAN, GauGAN, Deepfake, CRN, IMLE, Guided Diffusion, LDM, Glide, and Dalle. For synthetic detection experiments, both training and testing data are drawn exclusively from GenImage.

\smallskip
\noindent
\textbf{Implementation Details.} In the feature analysis stage, we randomly sample 6,000 real and 6,000 fake images from each GenImage subset, and 1,000–2,000 pairs per UniversalFakeDetect subset. AUC is computed per feature using Logistic Regression. For detection, we adopt the CLIP ViT-L/14 with LoRA (following C2pClip~\cite{tan2025c2p}), replacing only the captions with our core feature-based descriptions. Training uses Adam (lr=2e-4, batch size=128, 1 epoch) on 4× RTX 5090 GPUs, with images resized to 224×224.

\smallskip
\noindent
\textbf{Performance Evaluation Metric.}
For feature selection, we evaluate candidate features by analyzing their cross-dataset distributions, domain robustness, and classification performance (AUC and JMD values) to identify core discriminative features. In the testing phase, we use mean accuracy (mAcc) as the primary evaluation metrics to quantify synthetic detection performance.

\subsection{Results}

\smallskip
\noindent
\textbf{Feature Classification and Core Features.}
We extract 15 physical features from balanced real–fake image pairs and evaluate them using the Feature Stability and Discriminability Validity Analysis (FSDVA) framework. Based on stability (\(S_s\)) and discriminability (\(S_d\)) scores, we select four core features—LBP entropy, Laplacian variance, Sobel magnitude mean, and Sobel magnitude std—that satisfy both high stability and strong discriminability thresholds.

As shown in Fig.~\ref{fig:feat_cls}, the stability and discriminability scores of all 15 candidate features are visualized, with color-coded categories: green for core features (high \(S_s\) and \(S_d\)), orange for usable features, red for unstable but highly discriminative features, and grey for unusable features. This selection process ensures that the core features are robust to dataset variations while effectively distinguishing between real and fake images. Fig.~\ref{fig:core_feat} presents the density distributions of these four core features across five datasets (ADM, Midjourney, BigGAN, CRN, and Glide). We observe that the observed patterns are robust to dataset variations, validating the utility of these core features for downstream fake image detection.

\smallskip
\noindent
\textbf{Enhanced Captions with Core Features' Description.}
In Fig. \ref{fig:cap_samples} shows some samples of images and their corresponding semantic description and their enhanced captions with the core features' descriptions.

\smallskip
\noindent
\textbf{Results with different training data and compared with other methods.} Table \ref{tab:genimage_diff_train} presents a comparative evaluation of the generalization capabilities of C2pClip and our proposed method (Ours) across eight Genimage test sets, with both methods trained on varying source datasets. Our approach consistently outperforms C2pClip in most configurations, as indicated by the bolded values. In particular, when trained on the Wukong dataset, our method achieves near-perfect accuracy (e.g., 99.8\% on both SDv1.4 and Wukong test sets), underscoring its strong cross-model generalization. Conversely, C2pClip exhibits greater sensitivity to training data selection, with performance varying more substantially across different test domains. These findings demonstrate that our method not only adapts more effectively to diverse generative models but also maintains higher mean accuracy (mAcc) across nearly all training settings. Additionally, as shown in Table \ref{tab:genimage_acc}, our method achieves the highest average accuracy on the Genimage test sets overall.

\smallskip
\noindent
\textbf{Results Analysis.} To better understand the contribution of core features-based text and to eliminate potential confounding effects from training dynamics, network parameters, and hyperparameter settings, we conduct a controlled analysis using a pre-trained ViT-L/14 CLIP model. Specifically, we randomly sample 100 image-text pairs, each associated with three types of textual descriptions: (1) the original text generated by a pre-trained ClipCap model, (2) core features-based text derived from physical feature attributes, and (3) combined text that integrates both. For each pair, we extract image and text embeddings using CLIP and compute the cosine similarity. As shown in Fig. \ref{fig:Quantitative_evaluation}, the inclusion of core physical feature descriptions leads to higher cosine similarity scores for fake image-text pairs compared to their real counterparts, suggesting that feature-augmented text improves alignment with visual content in the shared embedding space. 

\section{Discussion}
\smallskip
\noindent
\textbf{Limitations.} Despite the promising results, this work has several limitations that warrant further investigation. First, due to the fixed token length constraint of the CLIP text encoder, we are unable to incorporate all available physical features with adaptive weighting, which limits the expressiveness of the input text. Moreover, it remains challenging to fully determine whether the injected physical features actively guide the model toward learning more generalizable representations or merely introduce statistical cues. Second, each physical attribute is currently represented as a single scalar value, which may oversimplify complex visual characteristics and potentially cause the model to overlook finer-grained local patterns that are critical for distinguishing real from fake images. Third, the interplay between the class prompt and the feature-based textual descriptions has not been systematically explored; understanding this interaction is essential to disentangle their respective contributions and avoid confounding effects. Finally, while our method demonstrates strong performance on the evaluated benchmarks, more comprehensive experiments across diverse generative models, datasets, and real-world scenarios are needed to fully establish its effectiveness and robustness.

\noindent
\textbf{Future Work.} Building on the above limitations, several directions for future work are worth pursuing. We plan to explore more flexible architectures, such as incorporating feature tokens directly into the vision encoder or leveraging cross-modal attention mechanisms, to bypass the token length limitations of CLIP and enable richer feature fusion. In addition, we aim to move beyond global feature aggregation by integrating localized physical descriptors—such as region-based texture or noise statistics—to capture discriminative spatial artifacts more effectively. Another important avenue is to systematically analyze the relationship between class prompts and feature-based descriptions, potentially through ablation studies or attention visualization, to better understand how they jointly influence model decisions. Furthermore, we intend to validate our approach on a broader range of generative models and unseen domains, including cross-lingual and multi-source synthetic data, to assess generalization and practical applicability. Ultimately, we hope to extend this framework toward more interpretable and controllable fake image detection, where physical feature cues can provide both performance gains and human-understandable justifications.

\section{Conclusion}
We are the first to propose the integration of physically grounded features into image caption generation, with the goal of guiding feature representation and enhancing semantic understanding for more generalizable fake image detection. To this end, we introduce a novel feature selection algorithm that identifies the most stable and discriminative physical attributes. Our experimental results demonstrate that captions enriched with physical feature descriptions significantly improve the model's ability to detect fake images. While the underlying mechanisms by which these features influence model behavior are not yet fully understood, our findings point to a promising direction for improving the generalization capability of vision-language models. Moreover, this approach opens up new possibilities for leveraging pixel-derived physical features to mitigate hallucinations and reduce errors introduced by inaccurate text captions in large language models.

\par\vfill\par


%
\section{Supplementary Materials}
\begin{figure}[t]
  \centering
  \includegraphics[width=1.0\textwidth]{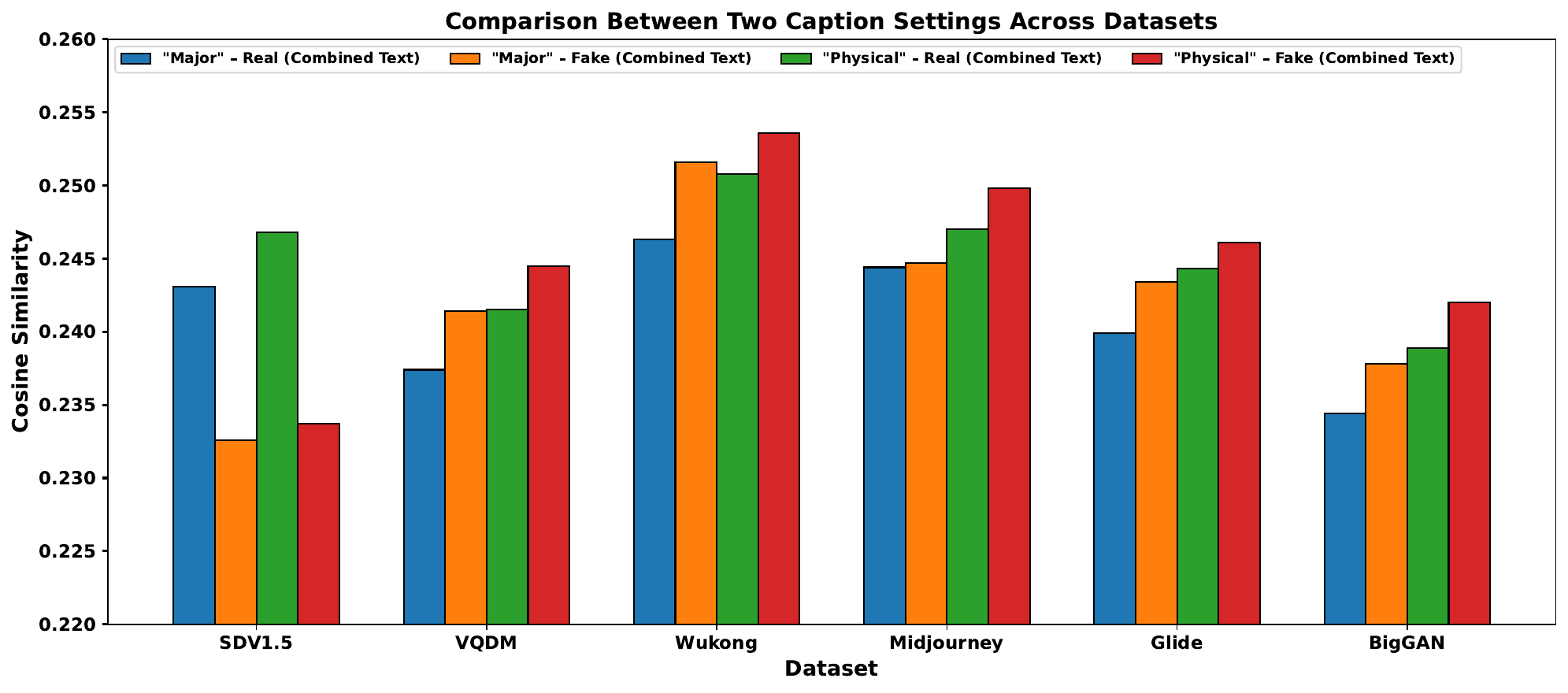}
  \caption{\textbf{Cosine similarity comparison between two caption styles across 7 datasets.} 
           Blue/orange: results using caption ``The major features are:''. 
           Green/red: results using caption ``The physical features are:''.}
  \label{fig:caption_comparison}
\end{figure}

\begin{figure}[!t]
  \centering
  \begin{minipage}{0.48\linewidth}
    \centering
    \includegraphics[width=\linewidth]{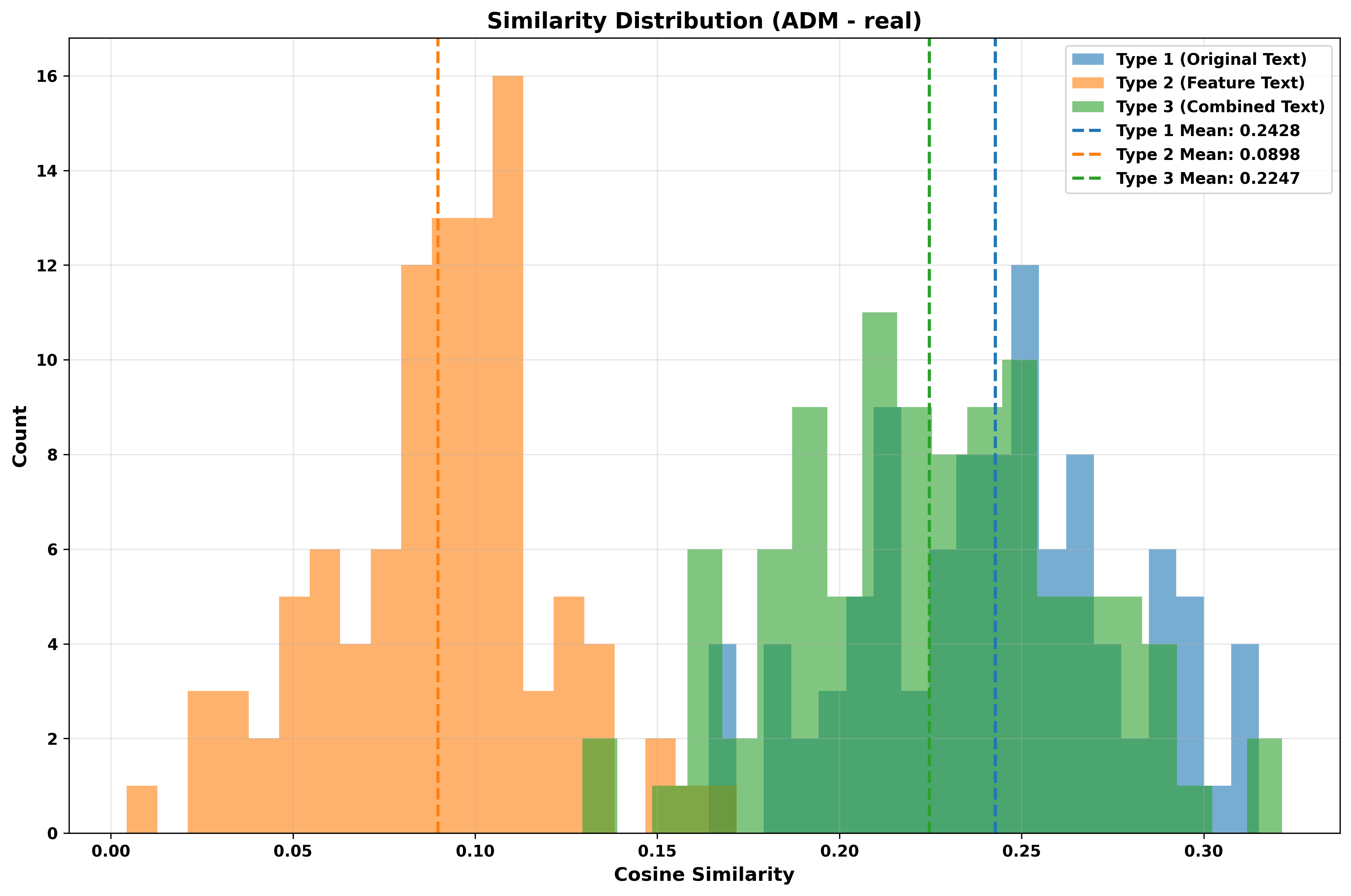}
    \label{fig:ADM_real}
  \end{minipage}
  \hfill
  \begin{minipage}{0.48\linewidth}
    \centering
    \includegraphics[width=\linewidth]{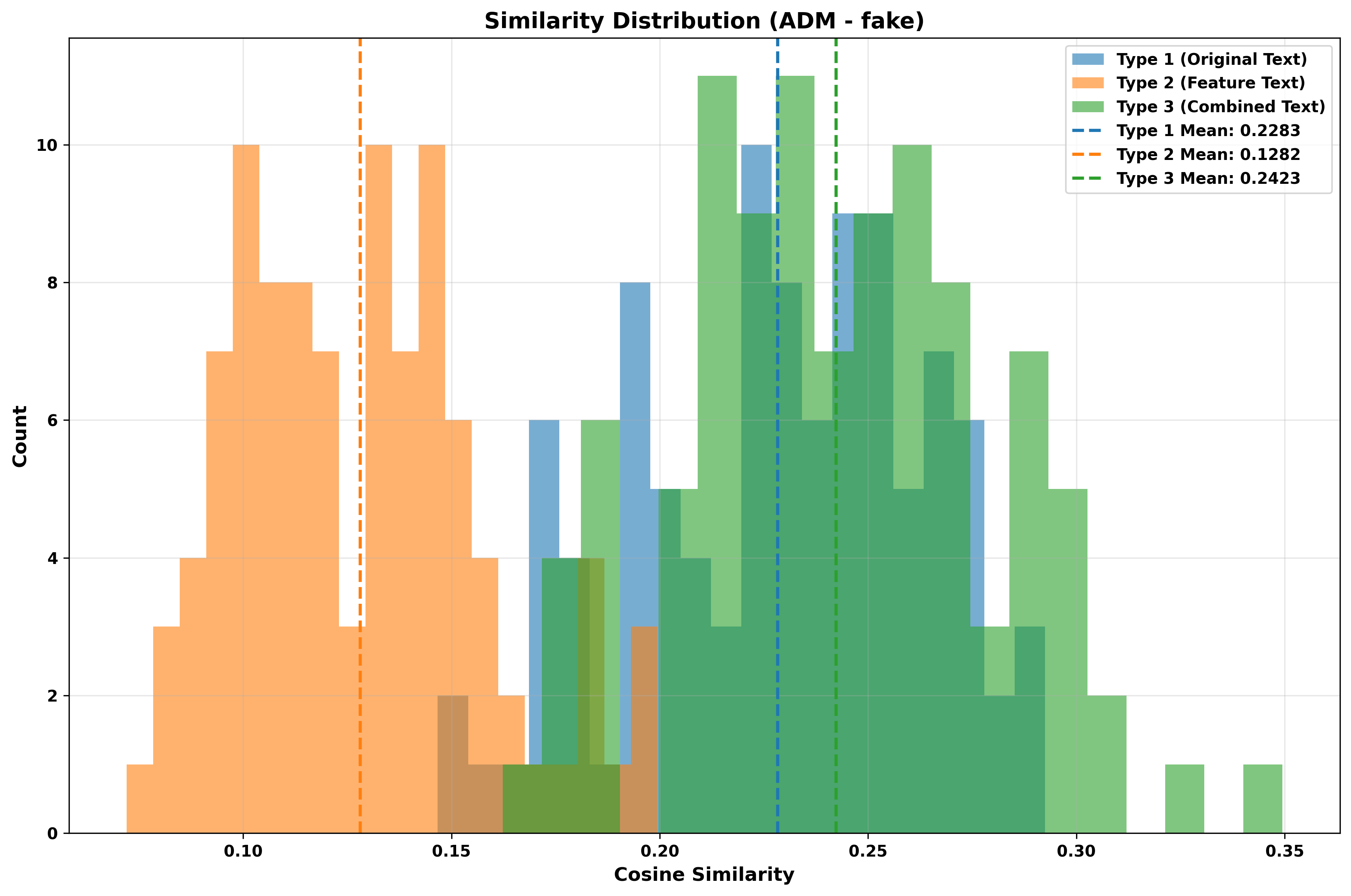}
    \label{fig:ADM_fake}
  \end{minipage}
  \caption{Image-text cosine similarity distributions of ADM}
  \label{fig:ADM}
\end{figure}

\begin{figure}[!t]
  \centering
  \begin{minipage}{0.48\linewidth}
    \centering
    \includegraphics[width=\linewidth]{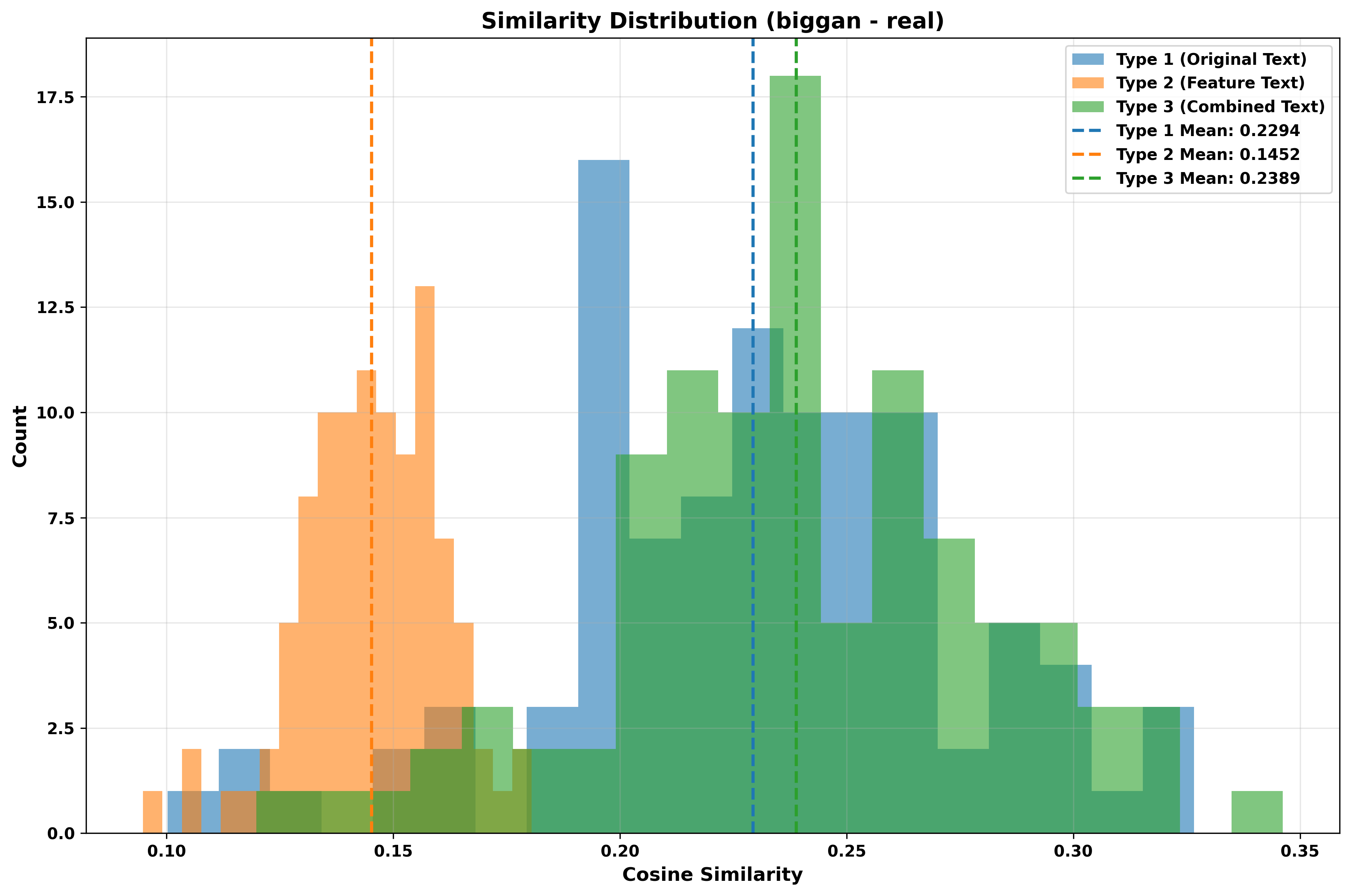}
  \end{minipage}
  \hfill
  \begin{minipage}{0.48\linewidth}
    \centering
    \includegraphics[width=\linewidth]{input_with_physicalfeatures/ADM_fake_similarity_analysis.png}
  \end{minipage}
  \caption{Image-text cosine similarity distributions of BigGAN}
  \label{fig:BigGAN}
\end{figure}

\begin{figure}[!t]
  \centering
  \begin{minipage}{0.48\linewidth}
    \centering
    \includegraphics[width=\linewidth]{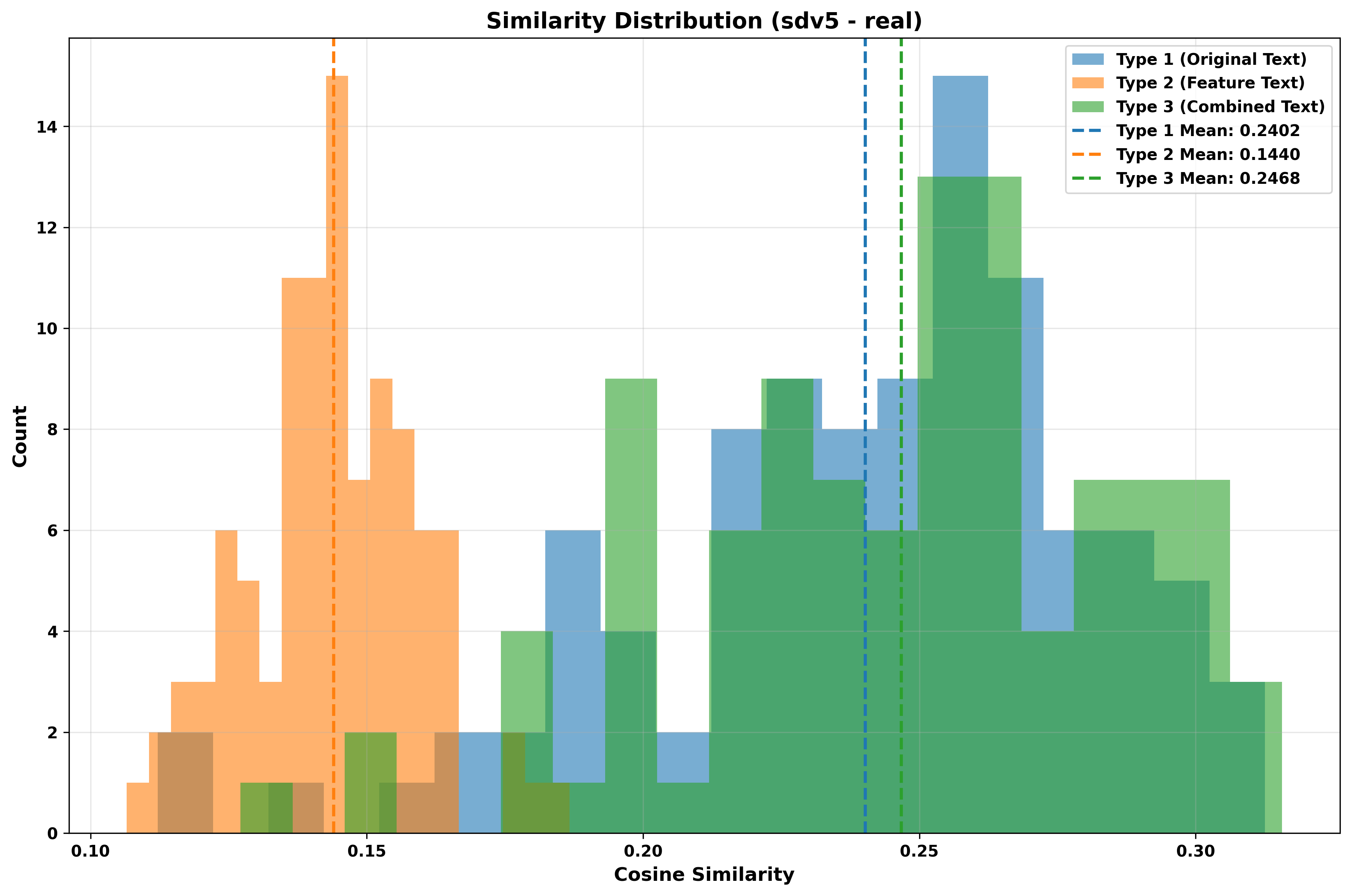}
  \end{minipage}
  \hfill
  \begin{minipage}{0.48\linewidth}
    \centering
    \includegraphics[width=\linewidth]{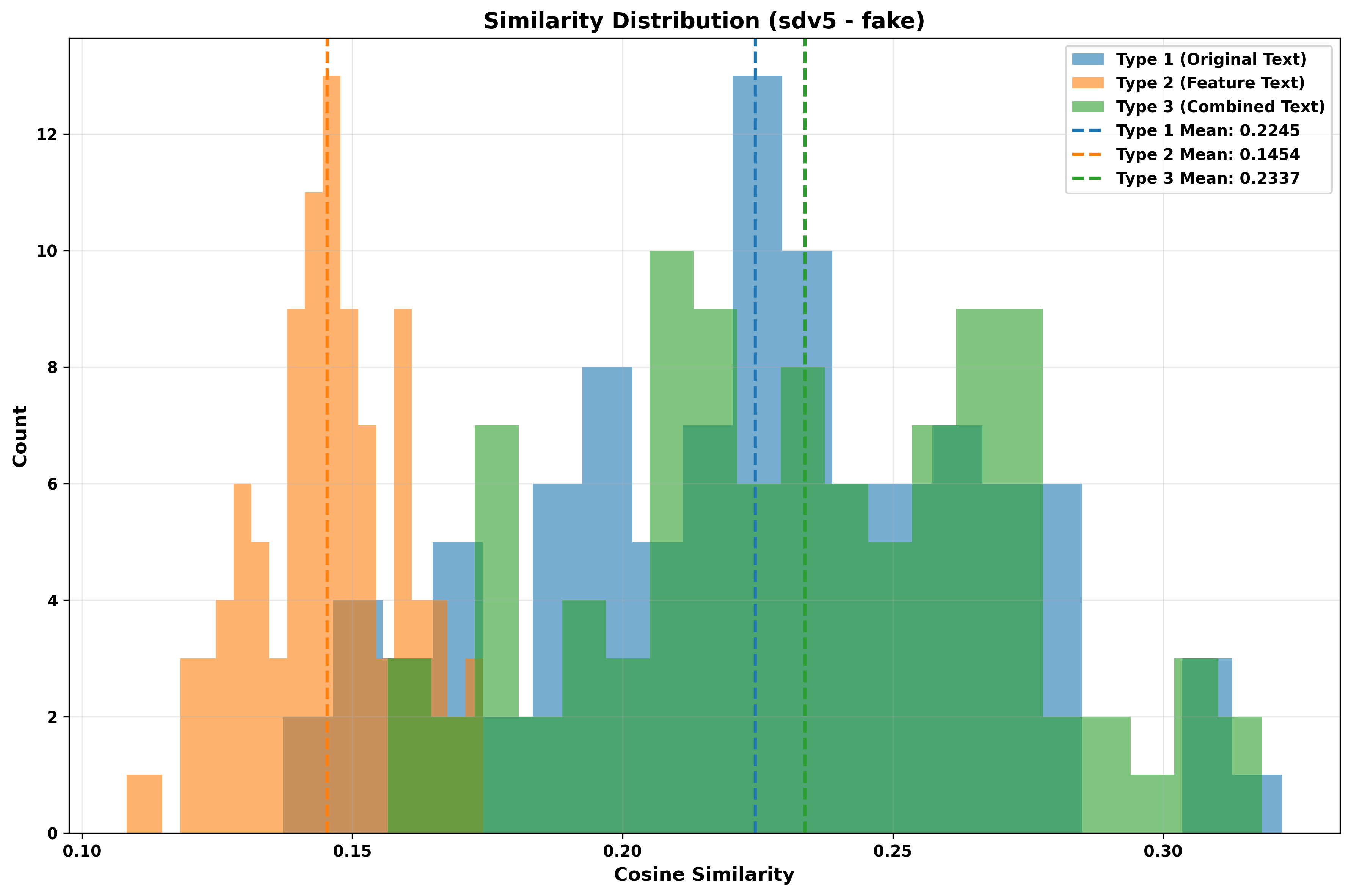}
  \end{minipage}
  \caption{Image-text cosine similarity distributions of SDv1.5}
  \label{fig:sdv5}
\end{figure}

\begin{figure}[!t]
  \centering
  \begin{minipage}{0.48\linewidth}
    \centering
    \includegraphics[width=\linewidth]{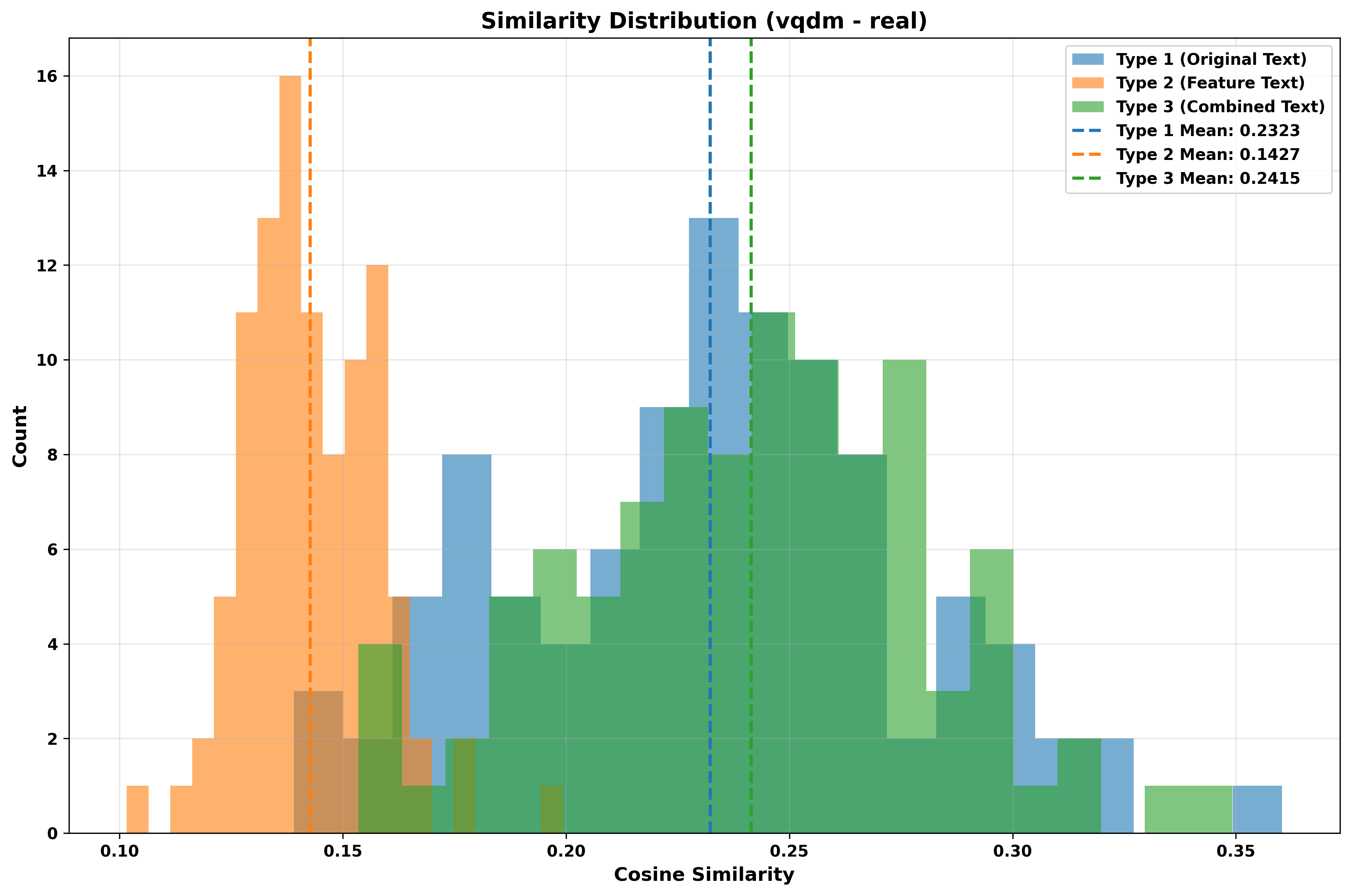}
  \end{minipage}
  \hfill
  \begin{minipage}{0.48\linewidth}
    \centering
    \includegraphics[width=\linewidth]{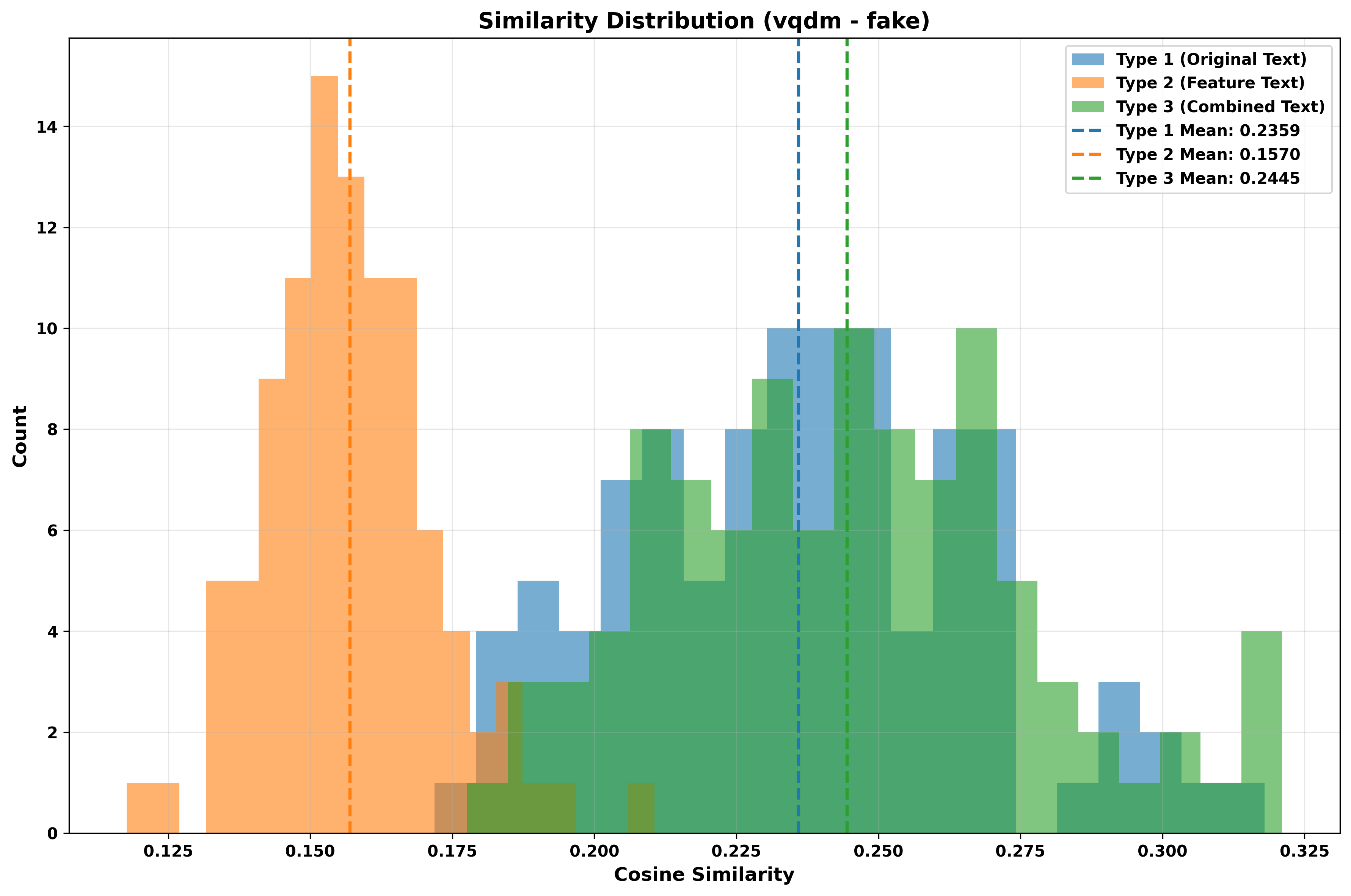}
  \end{minipage}
  \caption{Image-text cosine similarity distributions of VQDM}
  \label{fig:vqdm}
\end{figure}

\begin{figure}[!t]
  \centering
  \begin{minipage}{0.48\linewidth}
    \centering
    \includegraphics[width=\linewidth]{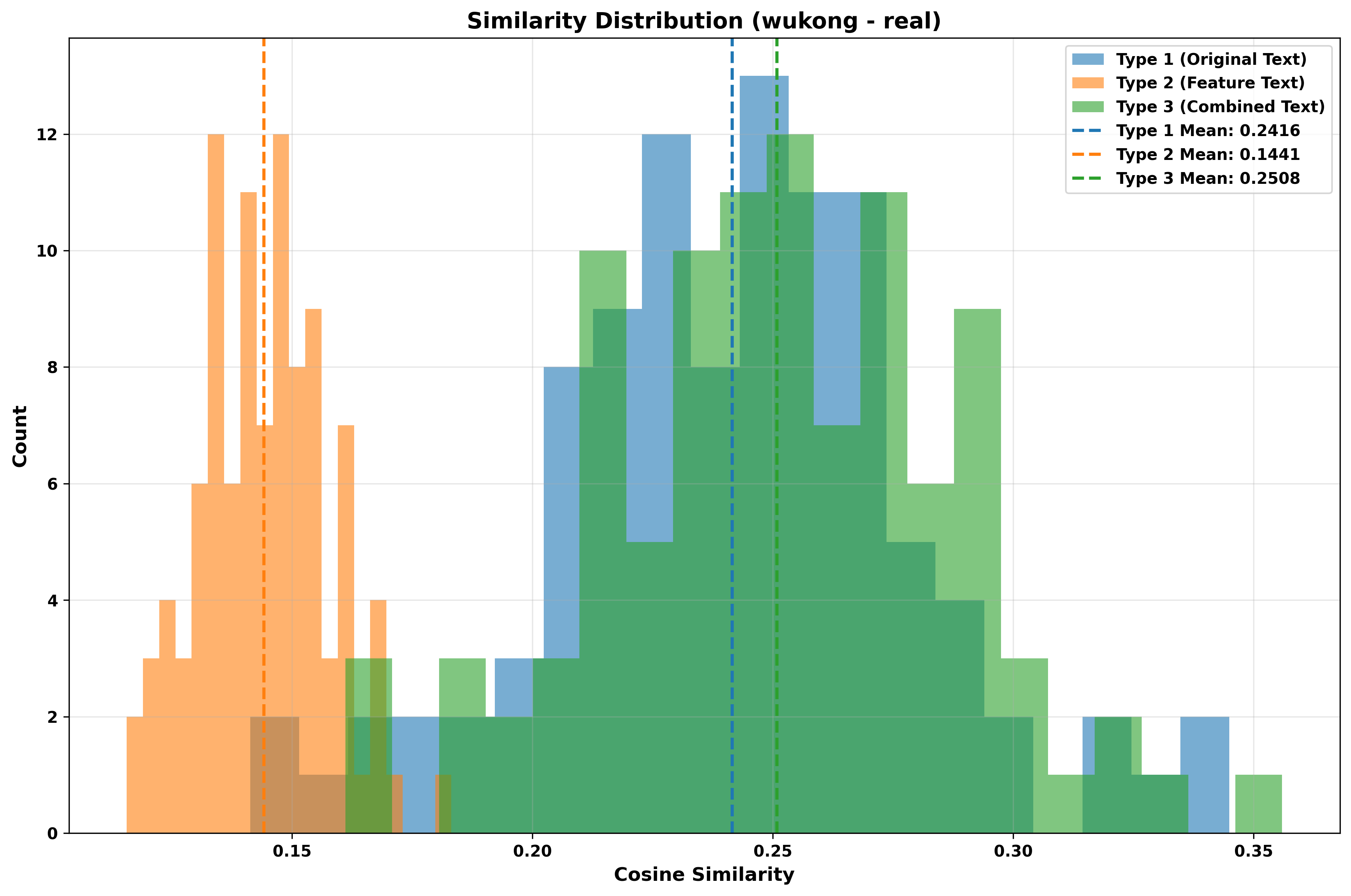}
    \label{fig:wukong_real}
  \end{minipage}
  \hfill
  \begin{minipage}{0.48\linewidth}
    \centering
    \includegraphics[width=\linewidth]{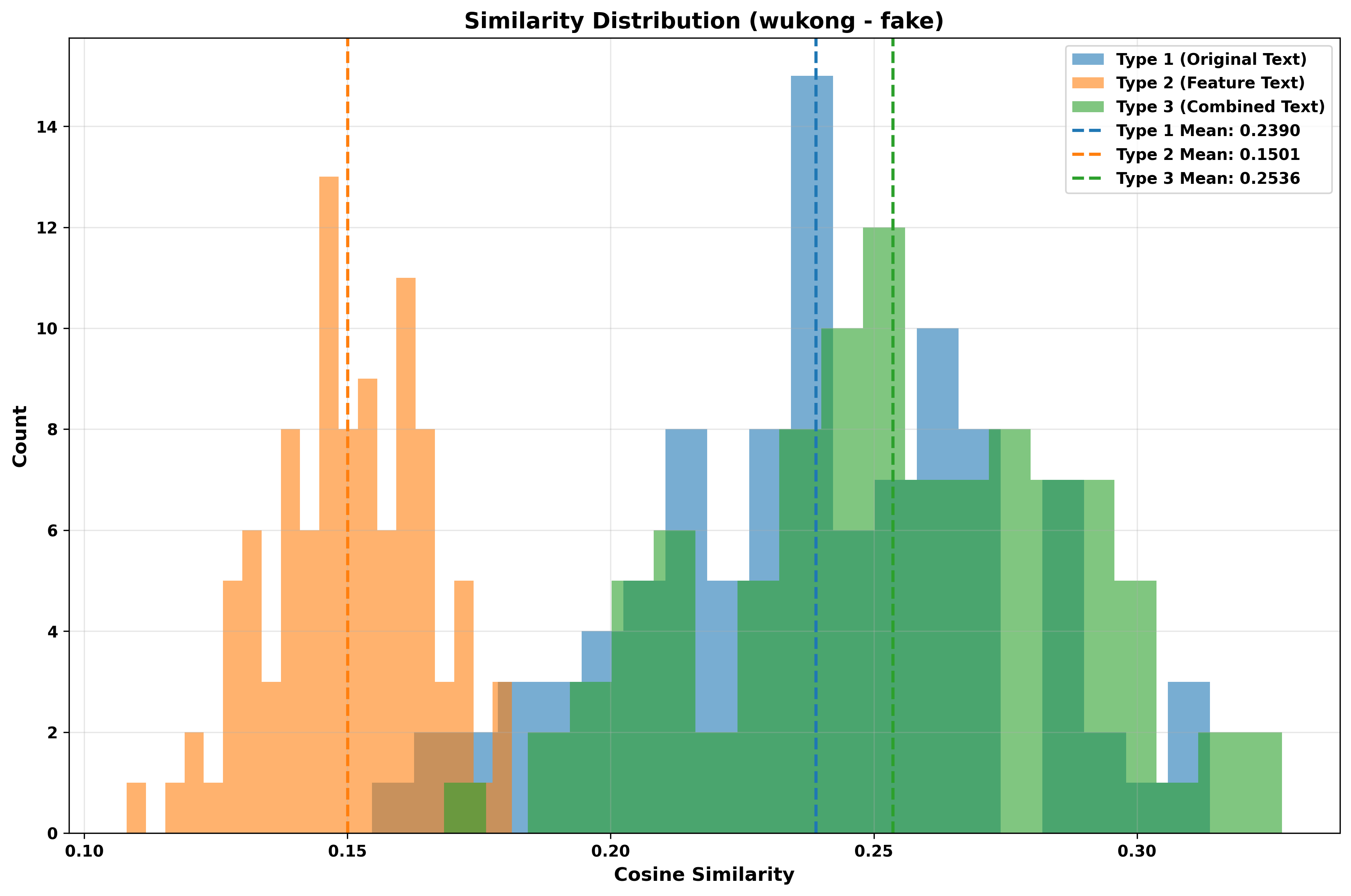}
    \label{fig:wukong_fake}
  \end{minipage}
  \caption{Image-text cosine similarity distributions of Wukong}
  \label{fig:wukong}
\end{figure}

\begin{figure}[!t]
  \centering
  \begin{minipage}{0.48\linewidth}
    \centering
    \includegraphics[width=\linewidth]{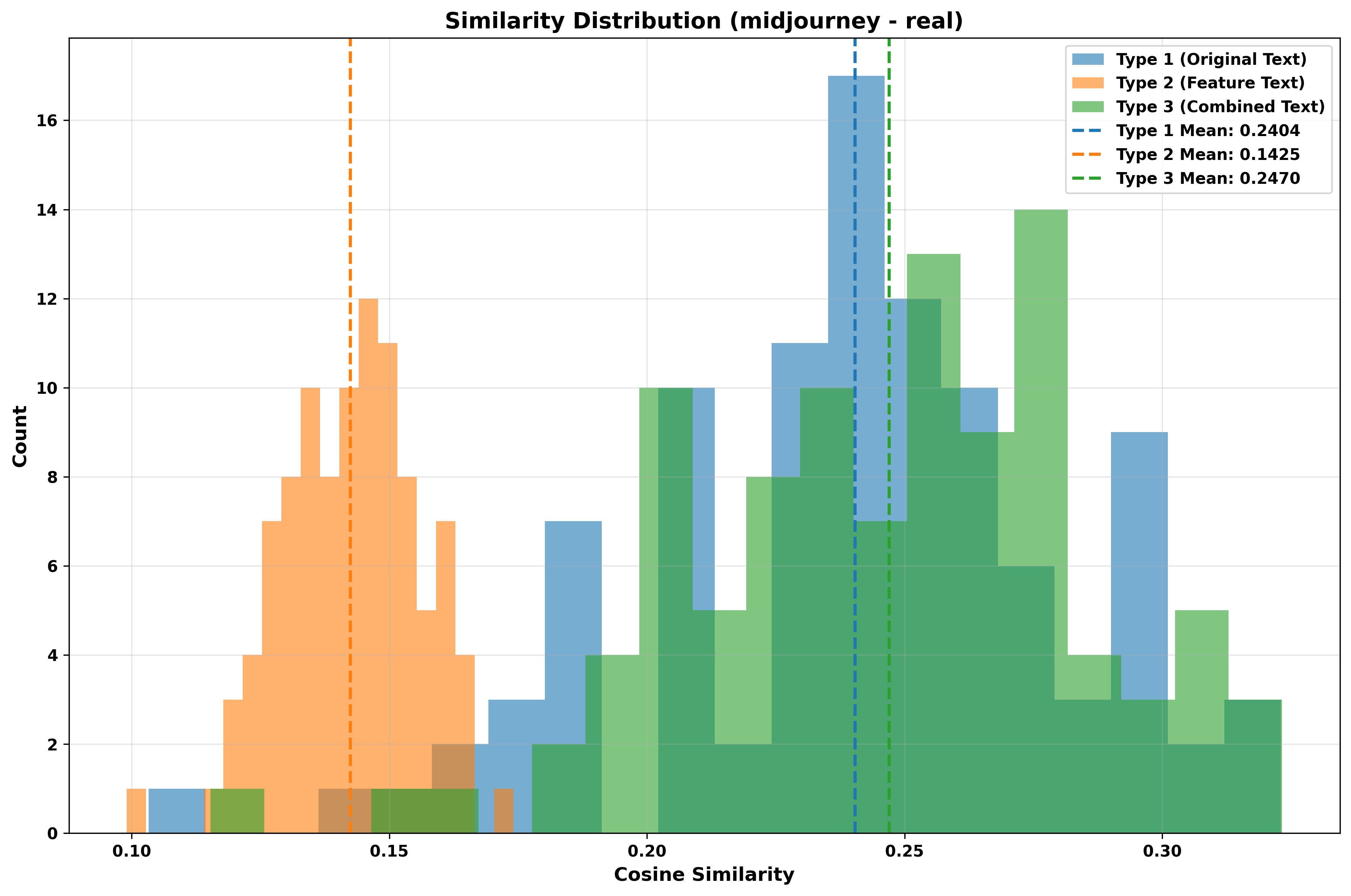}
    \label{fig:mid_real}
  \end{minipage}
  \hfill
  \begin{minipage}{0.48\linewidth}
    \centering
    \includegraphics[width=\linewidth]{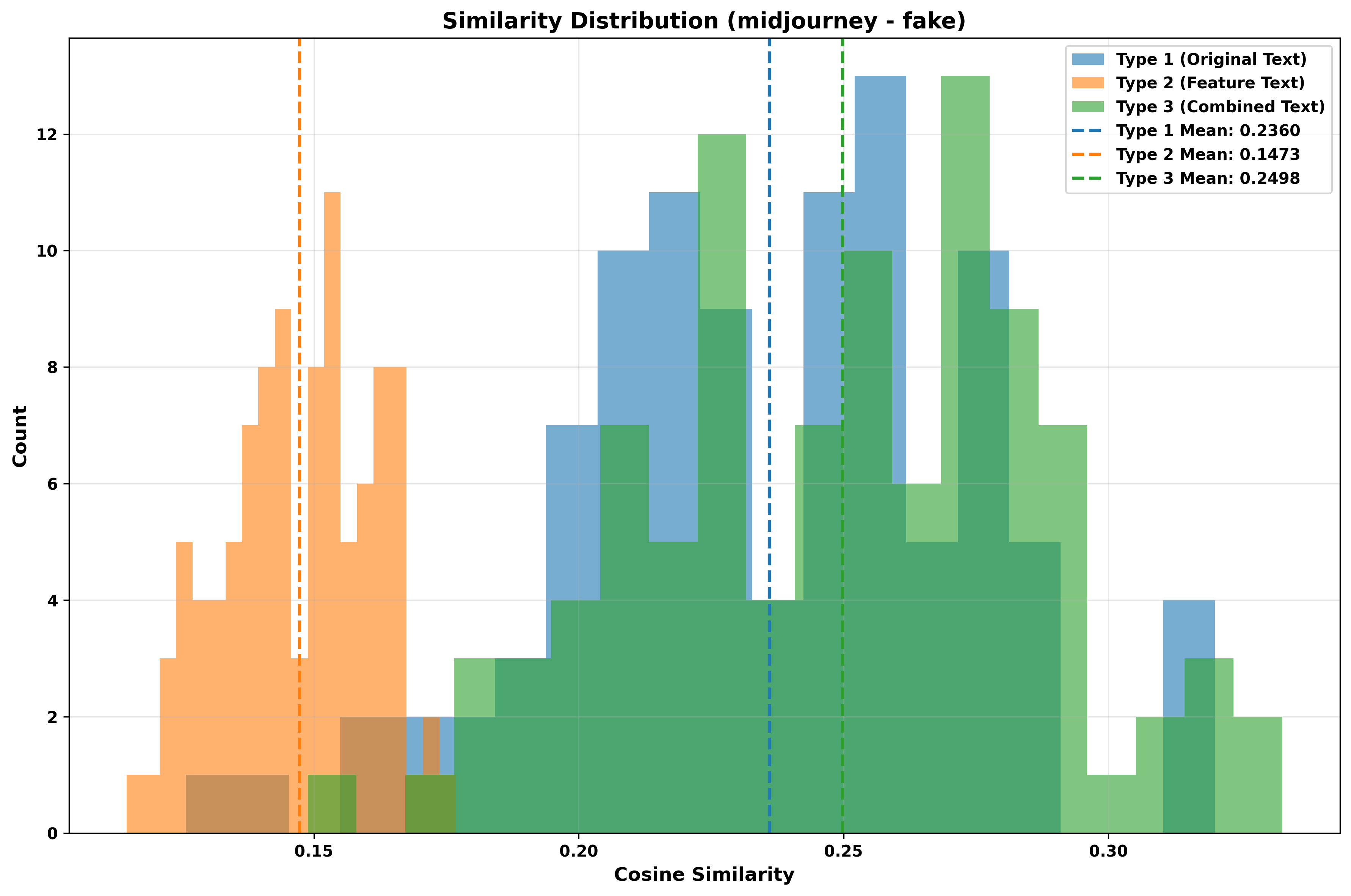}
    \label{fig:mid_fake}
  \end{minipage}
  \caption{Image-text cosine similarity distributions of Midjourney}
  \label{fig:mid}
\end{figure}

\begin{figure}[!t]
  \centering
  \begin{minipage}{0.48\linewidth}
    \centering
    \includegraphics[width=\linewidth]{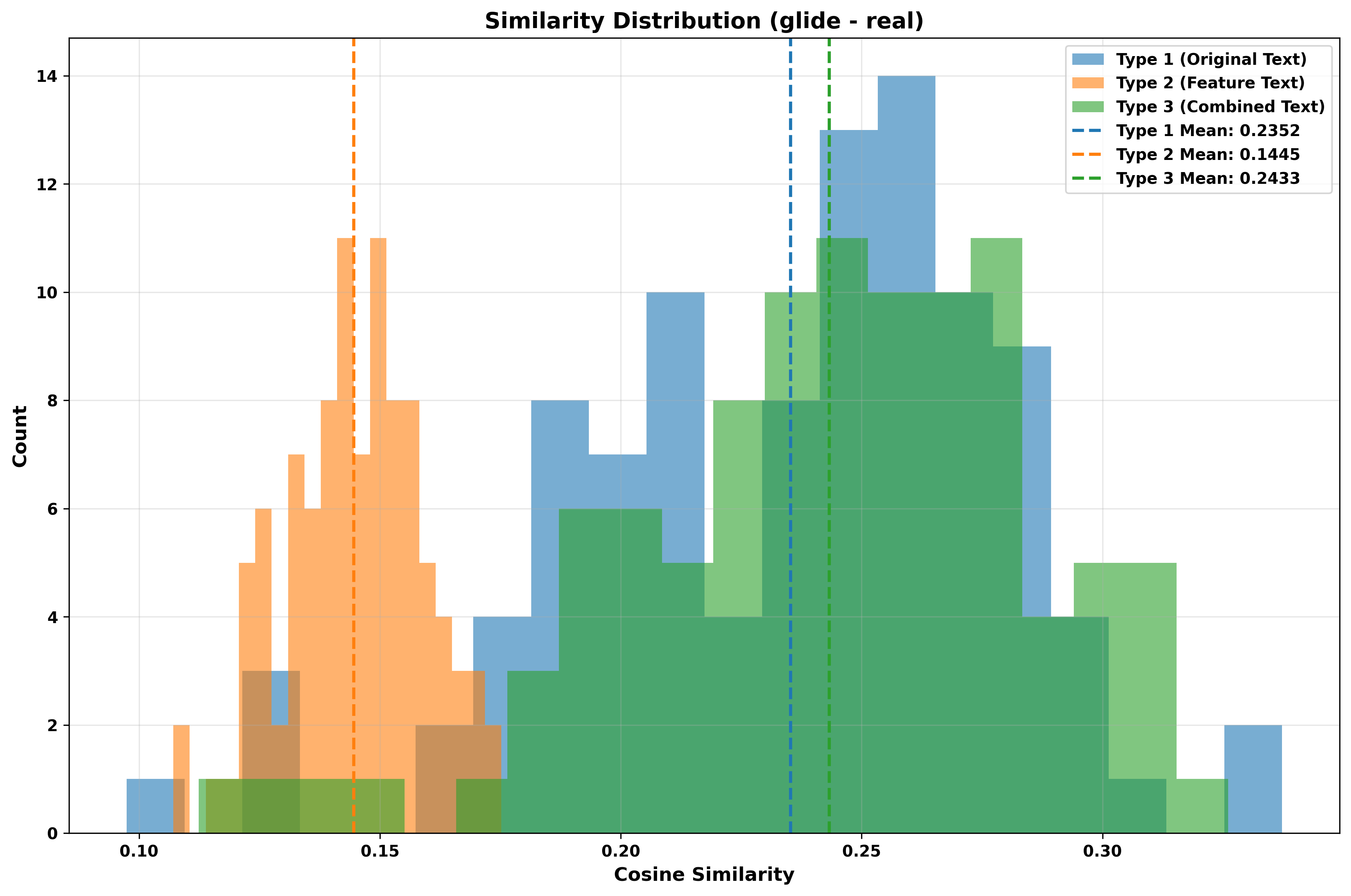}
    \label{fig:glide_real}
  \end{minipage}
  \hfill
  \begin{minipage}{0.48\linewidth}
    \centering
    \includegraphics[width=\linewidth]{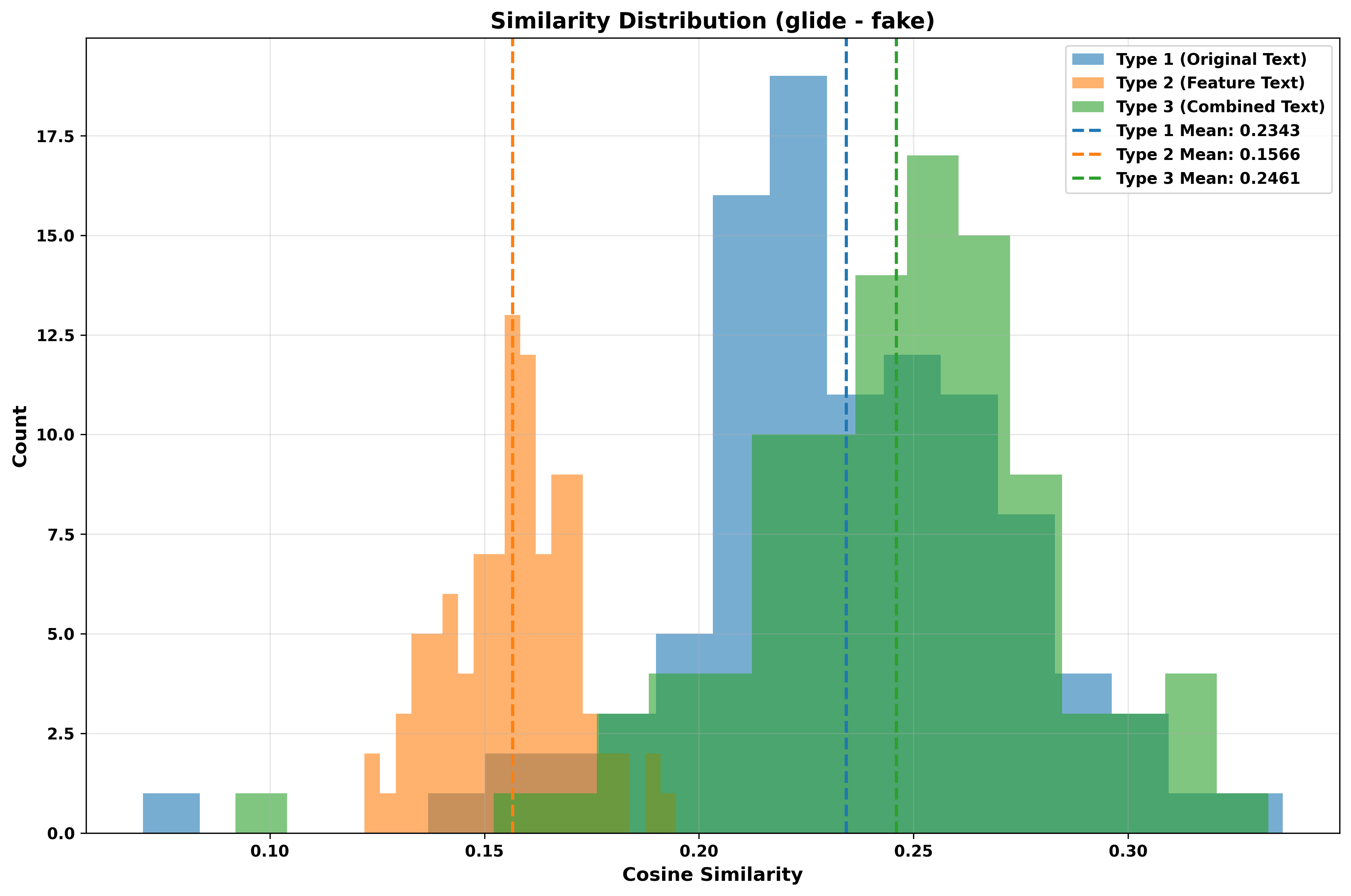}
    \label{fig:glide_fake}
  \end{minipage}
  \caption{Image-text cosine similarity distributions of Glide}
  \label{fig:glide}
\end{figure}

In this supplemental material, we provide a qualitative analysis of the impact of prompt phrasing on image–text alignment. Following the protocol described in Sec. 4.2, we compute cosine similarities between image embeddings and three types of textual descriptions using a frozen CLIP ViT-L/14 model: (1) original captions generated by ClipCap, (2) text consisting solely of core physical feature values extracted by our algorithm, and (3) combined text that integrates both. For the combined text, we initially used the introductory phrase “The major features are:” before listing the physical attributes. To assess the sensitivity of the alignment to linguistic framing, we replaced this phrase with “The physical features are:” while keeping the numerical feature values unchanged. This modification consistently increased the cosine similarity for both real and fake images across all evaluated samples, as shown in Fig. \ref{fig:caption_comparison}—suggesting that the revised phrasing better aligns the feature descriptions with the visual content in the shared embedding space. In Fig. \ref{fig:ADM} , Fig. \ref{fig:BigGAN}, Fig. \ref{fig:sdv5}, Fig. \ref{fig:vqdm} , Fig. \ref{fig:wukong},  Fig. \ref{fig:mid},  Fig. \ref{fig:glide}, we present the distribution of cosine similarity value of real and fake images from 7 of  the eight GenImage datasets (ADM, BigGAN, SDv1.5, Midjourney, GLIDE, Wukong, and VQDM) with the caption with “The physical features are:”. In each figure, we show the three caption types and their corresponding CLIP similarity scores, highlighting the improvement achieved by the refined prompt. These examples corroborate the quantitative trends and underscore the importance of linguistic precision when injecting physical descriptors into multimodal representations.



\bibliographystyle{splncs04}

\bibliography{main}
\end{document}